\documentclass[sigconf]{acmart}

\usepackage{booktabs} 
\usepackage{hyperref}
\usepackage{amsmath}
\usepackage{amsthm}
\usepackage[mathscr]{eucal}
\usepackage{stmaryrd}
\usepackage{algorithm}
\usepackage{algorithmic}
\usepackage{footmisc}   
\usepackage{subfig} 
\usepackage{setspace}   
\usepackage[utf8]{inputenc}
\usepackage[english]{babel}
\usepackage{xcolor}
\usepackage{graphicx}

\usepackage{enumerate}
\usepackage{accents}
\usepackage[framemethod=tikz]{mdframed}
\usepackage{colortbl}

\usepackage{tikz}
\usetikzlibrary{calc,fit,shapes,decorations.pathmorphing}

\graphicspath{{fig/}}   

\DeclareMathOperator*{\argmax}{argmax}
\newcommand{\eat}[1]{}

\newcommand{\bu}{\mathbf{u}}

\newcommand{\x}{\mathbf{x}}
\newcommand{\y}{\mathbf{y}}

\newcommand{\XCal}{\mathscr{X}}
\newcommand{\YCal}{\mathscr{Y}}
\newcommand{\PCal}{\mathscr{P}}

\newcommand{\OSf}{\mathsf{O}}
\newcommand{\SSf}{\mathsf{S}}
\newcommand{\defEq}{\stackrel{.}{=}}

\newcommand{\indicator}[1]{\llbracket #1 \rrbracket}

\newcommand\thickbar[1]{\accentset{\rule{.4em}{.8pt}}{#1}}

\newcommand{\eg}{e.g.\ }
\newcommand{\ie}{i.e.\ }

\allowdisplaybreaks

\begin{document}
\title{Revisiting revisits in trajectory recommendation}

\author{Aditya Krishna Menon, Dawei Chen, Lexing Xie, Cheng Soon Ong}
\affiliation{%
 \institution{The Australian National University \quad Data61/CSIRO}
}
\email{{ aditya.menon, u5708856, lexing.xie, chengsoon.ong }@anu.edu.au}

\begin{abstract}

Trajectory recommendation is the problem of recommending a sequence of places in a city for a tourist to visit.
It is strongly desirable for the recommended sequence to avoid loops, as tourists typically would not wish to revisit the same location.
Given some learned model that scores sequences, how can we then find the highest-scoring sequence that is loop-free?

This paper studies this problem, with three contributions.
First, we detail three distinct approaches to the problem -- 
graph-based heuristics,
integer linear programming, and
list extensions of the Viterbi algorithm
-- and qualitatively summarise their strengths and weaknesses.
Second, we explicate how two ostensibly different approaches to the list Viterbi algorithm are in fact fundamentally identical.
Third, we conduct experiments on real-world trajectory recommendation datasets to identify the tradeoffs imposed by each of the three approaches.

Overall, our results indicate that a greedy graph-based heuristic offer excellent performance and runtime, leading us to recommend its use for removing loops at prediction time.


\end{abstract}

%
%

\copyrightyear{2017}
\acmYear{2017}
\setcopyright{acmcopyright}
\acmConference{CitRec}{August 27, 2017}{Como, Italy}\acmPrice{15.00}\acmDOI{10.1145/3127325.3127326}
\acmISBN{978-1-4503-5370-0/17/08}

\ccsdesc[500]{Information systems~Recommender systems}

\maketitle

\section{Introduction}

A burgeoning sub-field of citizen-centric recommendation focusses on suggesting travel routes in a city that a tourist might enjoy.
This goal encompasses at least three distinct problems:
\begin{enumerate}[(1)]
	\item ranking \emph{all} points of interest (POIs) in a city in a manner personalised to a tourist (\eg a tourist to Sydney interested in scenic views might have {\tt Opera House $\succ$ Darling Harbour $\succ$ Chinatown}) \citep{shi2011personalized,lian2014geomf,hsieh2014mining,yuan2014graph};
	\item recommending the \emph{next} location a tourist might enjoy, given the sequence of places they have visited thus far (\eg given {\tt Darling Harbour$\to$Botanic Gardens}, we might recommend {\tt Quay Caf\'{e}}) \citep{fpmc10,ijcai13,zhang2015location};
	\item recommending an \emph{entire sequence} of POIs for a tourist, effectively giving them a travel itinerary (\eg {\tt Opera House$\to$Quay Caf\'{e}$\to$Darling Harbour}) \citep{lu2010photo2trip,ijcai15,lu2012personalized,gioniswsdm14,chen2015tripplanner}.
\end{enumerate}
Our focus in the present paper is problem setting (3), which we dub ``trajectory recommendation''.

Effectively tackling trajectory recommendation poses a challenge:
at training time, how can one design a model that can recommend sequences of POIs which are coherent as a \emph{whole}?
Merely concatenating a tourists' personalised top ranking POIs into a sequence
might result in prohibitive travel (e.g.\ {\tt Opera House$\to$Royal National Park}),
or unacceptably homogeneous results (e.g.\ we might recommend three restaurants in a row).
This motivates approaches that ensure \emph{global cohesion} of the predicted sequence.
In recent work, \citet{Chen:2017} showed that structured SVMs are one such viable approach which outperform POI ranking approaches. 

In this paper, we focus on a distinct but related challenge:
at prediction time, how can one recommend a sequence that does not have \emph{loops}?
This is desirable because tourists would typically wish to avoid revisiting a POI that has already been visited before.
In principle, this problem will not exist if one employs a suitably rich model which learns to suppress sequences with loops.
In practice, one is often forced to compromise on model richness owing to computational and sample complexity considerations.
We thus study this challenge, with the following contributions:
\begin{enumerate}
	\item[(\textbf{C1})] We detail three different approaches to the problem -- 
	graph-based heuristics,
	integer linear programming,
	and
	list extensions of the Viterbi algorithm
	-- and qualitatively summarise their strengths and weaknesses.
	\item[(\textbf{C2})] In the course of our analysis, we explicate how two ostensibly different approaches to the list Viterbi algorithm \citep{seshadri1994list,nilsson2001sequentially} are in fact fundamentally identical.
	\item[(\textbf{C3})] We conduct experiments on real-world trajectory recommendation datasets to identify the tradeoffs imposed by each of the three approaches.
\end{enumerate}
Overall, we find that
all methods offer performance improvements over na\"{i}vely predicting a sequence with loops,
but that
a greedy graph-based heuristic offers excellent performance and runtime.
We thus recommend its use
for removing loops at prediction time
over the more computationally demanding integer programming and list Viterbi algorithms. 


\section{Trajectory \& path recommendation}
\label{sec:background}

We now formalise the problem of interest and outline its challenges.

\subsection{Trajectory recommendation}

Fix some set $\PCal$ of points-of-interest (POIs) in a city.
A \emph{trajectory}\footnote{In graph theory, this is also referred to as a walk.} is any sequence of POIs, possibly containing loops (repeated POIs).
In the \emph{trajectory recommendation} problem, we are given as input a training set of historical tourists' trajectories.
From this, we wish to design a \emph{trajectory recommender}, which accepts a
\emph{trajectory query} $\x = (s, l)$, comprising a start POI $s \in \PCal$, and trip length $l \!>\! 1$, 
and produces one or more sequences of $l$ POIs starting from $s$. 

Formally, let $\XCal \defEq \PCal \times \{ 2, 3, \ldots \}$ be the set of possible queries,
$\YCal \defEq \bigcup_{l = 2}^\infty \PCal^l$ be the set of all possible trajectories,
and for fixed $\x \in \XCal$, $\YCal_{\x} \subset \YCal$ be the set of trajectories that conform to the constraints imposed by $\x$,
\ie if $\x = (s, l)$ then
$\YCal_{\x} = \{ \y \in \PCal^l \mid y_1 = s \}$.
Then, the {trajectory recommendation} problem has:

\vspace{0.5\baselineskip}

\begin{mdframed}[innertopmargin=3pt,innerbottommargin=3pt,skipbelow=5pt,roundcorner=8pt,backgroundcolor=red!3,topline=false,rightline=false,leftline=false,bottomline=false]
	\begin{tabular}{ll}
		{\sc Input}:  & training set $\left\{ \left( \x^{(i)}, \y^{(i)} \right) \right\}_{i = 1}^n \in ( \XCal \times \YCal )^n$ \\
		{\sc Output}: & a trajectory recommender $r \colon \XCal \to \YCal$ \\
	\end{tabular}
\end{mdframed}

\vspace{0.5\baselineskip}

One way to design a trajectory recommender is to find a (query, trajectory) affinity function $f \colon \XCal \times \YCal \to \mathbb{R}$, and let
\begin{equation}
	\label{eqn:argmax}
	r( x ) \defEq \argmax_{\y \in \YCal_x}~f(\x, \y).
\end{equation}
Several choices of $f$ are possible.
\citet{cikm16paper} proposed to use $f$ given by a RankSVM model. 
While offering strong performance, this has a conceptual disadvantage highlighted in the previous section:
it does not model global cohesion, and could result in solutions such as recommending three restaurants in a row.

To overcome this, \citet{Chen:2017} proposed to use $f$ given by a structured SVM (SSVM),
wherein $f( \x, \y ) = \mathbf{w}^T \Phi( \x, \y )$ for a suitable feature mapping $\Phi$.
When this feature mapping decomposes into terms that depend only on adjacent elements in the sequence $\y$ (akin to a linear-chain conditional random field),
the optimisation in Equation \ref{eqn:argmax} can be solved with the classic Viterbi algorithm.


\tikzstyle{state}=[shape=circle,draw=blue!50,fill=blue!20]
\tikzstyle{state2}=[shape=circle,draw=purple!50,fill=purple!20]
\tikzstyle{hiddenState}=[shape=circle,draw=gray!50,fill=gray!20,dashed]
\tikzstyle{specialState}=[shape=circle,double=red,draw=blue!50,fill=blue!20,dashed]
\tikzstyle{observation}=[shape=rectangle,draw=orange!50,fill=orange!20]
\tikzstyle{hiddenObservation}=[shape=rectangle,draw=gray!50,fill=gray!20,dashed]
\tikzstyle{lightedge}=[<-,thin]
\tikzstyle{mainstate}=[state,ultra thick]
\tikzstyle{mainedge}=[<-,ultra thick]

\begin{figure*}[!htb]
    \centering
    \subfloat[{\sc LoopElim} (\S\ref{sec:loop-elim}).]{
    \begin{tikzpicture}[baseline=(s0.base)]
        \node[specialState] (s0) at (0,0) {$1$};
        \node[specialState] (s1) at (1,0) {$2$}
            edge [<-,ultra thick] (s0);
        \node[specialState] (s2) at (2,0) {$3$}
            edge [<-,ultra thick] (s1);
        
        \draw [<-,ultra thin,bend right] (s0) to [looseness=1.25] (s2) node[sloped,draw=none] at (1,-0.45) {$/$};

        \node[draw=none] at (0,-1.15)  {};
    \end{tikzpicture}
    }%
    \qquad    
    \subfloat[{\sc Greedy} (\S\ref{sec:greedy}).]{
    \begin{tikzpicture}[baseline=(s12.base)]
        \node[specialState] (s11) at (0,1)  {$1$};
        \node[hiddenState]  (s12) at (0,0)  {$2$};
        \node[hiddenState]  (s13) at (0,-1) {$3$};

        \node[hiddenState]  (s21) at (1,1)  {$1$};
        \node[specialState] (s22) at (1,0)  {$2$};
        \node[hiddenState]  (s23) at (1,-1) {$3$};

        \node[draw=none] (s32) at (2,1)  {};
        \node[draw=none] (s31) at (2,0)  {};
        \node[draw=none] (s33) at (2,-1) {};

        \node[draw=none] (s411) at (2.25,0)     {$\ldots$};

        \draw [->,ultra thin]      (s11) to (s21) node [draw=none] at (0.5,1) {$/$};
        \draw [->,ultra thick] (s11) to (s22);
        \draw [->,ultra thin]  (s11) to (s23);

        \draw [->,ultra thin] (s22) to (s32) node [draw=none] at (1.5,0.5) {$/$};
        \draw [->,ultra thin] (s22) to (s31) node [draw=none] at (1.5,0)     {$/$};
        \draw [->,ultra thin] (s22) to (s33);
    \end{tikzpicture}
    }%
    \qquad
    \subfloat[{\sc ILP} (\S\ref{sec:ilp}).]{
    \begin{tikzpicture}[baseline=(s1.base)]
        \node[specialState] (s1) at (0,0)  {$1$};
        \node[specialState] (s2) at (1,1)  {$2$};
        \node[hiddenState]  (s3) at (2,1)  {$3$};
        \node[hiddenState]  (s4) at (1,-1) {$4$};
        \node[specialState] (s5) at (2,-1) {$5$};
        \node[specialState] (s6) at (3,0)  {$6$};


        \draw [->,ultra thick] (s1) to (s2);
        \draw [->,ultra thin] (s1) to (s3);
        \draw [->,ultra thin] (s1) to (s4);
        \draw [->,ultra thin] (s1) to (s5);
        \draw [->,ultra thin] (s1) to (s6);    

        \draw [->,ultra thin] (s2) to (s1);
        \draw [->,ultra thin] (s2) to (s3);
        \draw [->,ultra thin] (s2) to (s4);
        \draw [->,ultra thick] (s2) to (s5);
        \draw [->,ultra thin] (s2) to (s6);    

        \draw [->,ultra thin] (s3) to (s1);
        \draw [->,ultra thin] (s3) to (s2);
        \draw [->,ultra thin] (s3) to (s4);
        \draw [->,ultra thin] (s3) to (s5);
        \draw [->,ultra thin] (s3) to (s6);    

        \draw [->,ultra thin] (s4) to (s2);
        \draw [->,ultra thin] (s4) to (s3);
        \draw [->,ultra thin] (s4) to (s1);
        \draw [->,ultra thin] (s4) to (s5);
        \draw [->,ultra thin] (s4) to (s6);    

        \draw [->,ultra thin] (s5) to (s2);
        \draw [->,ultra thin] (s5) to (s3);
        \draw [->,ultra thin] (s5) to (s4);
        \draw [->,ultra thin] (s5) to (s1);
        \draw [->,ultra thick] (s5) to (s6);    

        \draw [->,ultra thin] (s6) to (s2);
        \draw [->,ultra thin] (s6) to (s3);
        \draw [->,ultra thin] (s6) to (s4);
        \draw [->,ultra thin] (s6) to (s5);
        \draw [->,ultra thin] (s6) to (s1);                    
    \end{tikzpicture}
    }%
    \qquad
    \subfloat[{\sc List Viterbi} (\S\ref{sec:viterbi}).]{
    \begin{tikzpicture}[baseline=(s1.base)]
        \node[specialState] (s1) at (0,0)  {$1$};
        \node[specialState] (s2) at (1,0)  {$2$}            
            edge [<-,ultra thick] (s1);

        \node[specialState] (ss1) at (2,-1) {{${5}$}}
        	edge [<-,ultra thick,decorate,decoration={snake}] (s2);
        \node[specialState] (ss2) at (3,-1) {{${6}$}}
            edge [<-,ultra thick,decorate,decoration={snake}] (ss1);

        \node[hiddenState] (s3) at (2,0)  {$3$}            
            edge [<-,ultra thin] (s2);


        \draw [<-,ultra thin,bend left] (s1) to [looseness=1.25] (s3); 
    \end{tikzpicture}
    }
    
    \caption{Schematics of different algorithms to return a loop-free prediction.
    Nodes such as
	{\protect\tikz[baseline=(X.base)]{\protect\node[specialState,inner sep=2pt] (X) {\small$1$}}}
    are selected by the algorithm, with thick edges such as
    {\protect\tikz{\protect\coordinate (X) at (0,0) {}; \protect\coordinate (Y) at (0.5,0) {}; \protect\draw[->,ultra thick] (X) to (Y); \protect\node[draw=none] at (0.25,-0.00625) {}}}
    denoting the sequence ordering.
    {\sc LoopElim} removes the loop from the Viterbi solution (here the POI sequence $(1,2,3,1)$), possibly returning a path of shorter length than requested;    
    {\sc Greedy} incrementally selects POIs which have not been selected before, and locally maximises the sub-path score;
    {\sc ILP} solves an integer linear program to find the optimal length $l$ path in a complete graph over POIs;
    {\sc ListViterbi} finds where the second-best sequence diverges from the standard Viterbi sequence ($(1,2,3,1)$ as before); if not loop-free, it finds where the third-best diverges from the second-best, \emph{etc}.
    }
    \label{fig:schematics}
    \vspace{-0.5\baselineskip}
\end{figure*}
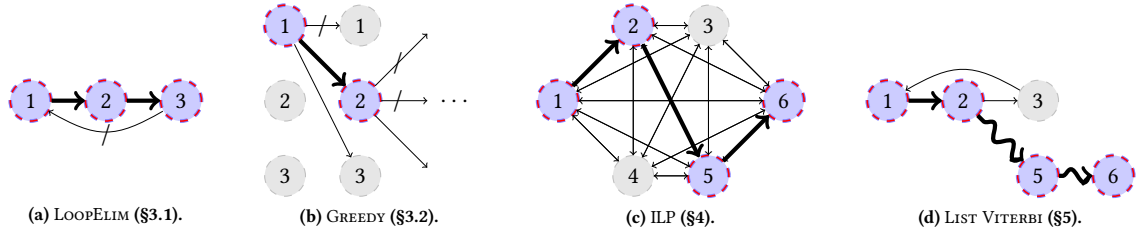

\subsection{Path recommendation}

We argue that the definition of trajectory recommendation is incomplete for a simple reason:
in most cases, a tourist will not want to revisit the same POI.
Instead, what is needed is to recommend a \emph{path}, \ie a trajectory that does not have any repeated POIs.
Let $\thickbar{\YCal} \subset \YCal$ be the set of all possible paths,
and for fixed $\x \in \XCal$, let $\thickbar{\YCal}_{\x} \subset \thickbar{\YCal}$ be the set of paths that conform to the constraints imposed by $\x$.
We now wish to construct a \emph{path recommender} $r \colon \XCal \to \thickbar{\YCal}$ via
\begin{equation}
	\label{eqn:argmax-path}
	r( x ) \defEq \argmax_{\y \in \thickbar{\YCal}_x}~f(\x, \y).
\end{equation}
For $f$ given by an SSVM, Equation \ref{eqn:argmax-path} requires we depart from the standard Viterbi algorithm, as the sequence in Equation \ref{eqn:argmax} may well have a loop.\footnote{In SSVMs, this issue also arises during training \citep{Chen:2017}, but we focus here only on the prediction problem. See \S\ref{sec:discussion} for additional comments.}
There are two distinct modes of attack available:
\begin{enumerate}
	\item seek an approximate solution to the problem,
	via heuristics that exploit a graph view of Equation \ref{eqn:argmax-path}.

	\item seek an exact solution to the problem,
	via integer linear programming,
	or top-$K$ extensions of the Viterbi algorithm. 
\end{enumerate}
While \citet{Chen:2017} suggested the latter exact approaches, they did not formally compare their performance either qualitatively or quantitatively;
they did not detail the different top-$K$ extensions of the Viterbi algorithm and the connections thereof;
and they did not consider approximate approaches.

In the sequel, we thus detail the above approaches in more detail.
Figure \ref{fig:schematics} gives a schematic overview of these algorithms.


\section{Graph-based heuristics}
\label{sec:heuristics}

To design approximations, it will be useful to view Equation \ref{eqn:argmax-path} as a graph optimisation problem
for the case of structured SVMs with pairwise potentials.
Here, the score $f(\mathbf{x}, \mathbf{y})$ can be decomposed as
\begin{equation}
	\label{eqn:f-ssvm}
	f(\mathbf{x}, \mathbf{y}) = \sum_{k = 1}^{|\mathbf{y}|} \alpha( y_k ) + \sum_{k = 1}^{|\mathbf{y}| - 1} \beta( y_k, y_{k+1} )
\end{equation}
for suitable unary and pairwise potentials $\alpha, \beta$~\cite{Chen:2017}.
Consider then a complete graph where the nodes are POIs, each node $p \in \PCal$ has a score $\alpha( p )$, and each edge $(p, p') \in \PCal^2$ has a score $\beta( p, p' )$.
Then, Equation \ref{eqn:argmax-path} is equivalently a problem of selecting $l$ nodes whose
total sum of node and edge scores maximises Equation \ref{eqn:f-ssvm}.
We now look at ways of approximately solving this selection problem.

\subsection{Heuristic loop elimination}
\label{sec:loop-elim}

Perhaps the simplest approximate solution to Equation \ref{eqn:argmax-path} is to merely remove the first loop occurring in the standard Viterbi solution (Equation \ref{eqn:argmax}).
That is, if the Viterbi solution is $( y_1, \ldots, y_l )$,
return the sub-sequence $( y_1, \ldots, y_{i-1} )$
for the first index $i$ where $y_i$ appears in this sub-sequence.
This has complexity dominated by the Viterbi algorithm, \emph{viz.} $\mathscr{O}(l \cdot |\PCal|^2)$ for an input query $\x = (s, l)$.

From a graph perspective, this approach makes the directed sub-graph induced by the Viterbi solution acyclic
by breaking the first cycle-inducing edge.
This is sensible if sequences never escape the first cycle, \ie after the first repeated POI, there is no new POI.
We have indeed found this to be the case in our problem; 
more generally, 
the problem of removing cycles is a special case of the ({\sf NP}-hard) minimum feedback arc-set problem \citep[pg.\ 192]{Garey:1990}.

This algorithm is appealing in its simplicity,
but has at least two drawbacks.
First, it makes the questionable assumption that Equation \ref{eqn:argmax-path} is solvable from the standard Viterbi solution alone.
Second, the solution violates the length constraint of the path recommendation problem.
As a remedy,
we can request the Viterbi algorithm to return a sequence of longer length
$l' \in \{ l, l + 1, \ldots, |\PCal| \}$, and pick the (smallest) $l'$ for which the predicted length is closest to $l$.


\subsection{Greedy path discovery}
\label{sec:greedy}

In light of the graph-based view, a natural approach to approximately solve Equation \ref{eqn:argmax-path} is a greedy algorithm.
Suppose we have already determined a partial path comprising distinct POIs $y_1, \ldots, y_k$.
Then, we can select the next candidate POI $y_{k + 1}$ as
the node
that iteratively optimises Equation \ref{eqn:f-ssvm},
subject to the constraint that it is distinct from all other nodes in the current path;
formally, we pick
$$ y_{k + 1} = \argmax_{p \in \PCal - \{ y_1, \ldots, y_k \}} \alpha( p ) + \beta( y_k, p ). $$
This algorithm is faster than the above heuristic, with $\mathscr{O}(l \cdot | \PCal |)$ complexity,
but similarly has an unclear performance guarantee.


\section{Integer Linear Programming \& TSP}
\label{sec:ilp}




One way to exactly solve Equation \ref{eqn:argmax-path}
is to observe its similarity to the travelling salesman problem (TSP).
Indeed, if the length $l = |\PCal|$, so that every POI is restricted to be visited once,
then we exactly get the TSP problem;
however, for smaller $l$, we require visiting only a \emph{subset} of POIs,
which is not a vanilla instance of TSP.

Nonetheless, we may take inspiration from methods to solve the TSP to attack our problem.
In particular, 
we can formulate the trip recommendation problem as an \emph{integer linear program} (\emph{ILP}),
as often done for the TSP \citep{opt98}.
Formally, given a starting location $s$ and the required trip length $l$,
we find the best possible path
via \citep{Chen:2017}

\resizebox{0.9625\linewidth}{!}{
\begin{minipage}{0.9625\linewidth}
	\begin{align*}
	\max_{\bu, \mathbf{v}} & \sum_{k=1}^m \alpha( p_k ) \cdot \sum_{j=1}^m u_{jk} +
	            \sum_{j,k=1}^m u_{jk} \cdot \beta( p_j, p_k ) \\
	s.t. 
	& \sum_{k=2}^m u_{1k} = 1, \, z(\bu)_1 = 0, z(\bu)_i \in \{ 0, 1 \} \, &&(\forall i \in \{2,\cdots,m\}) \\
	& \sum_{j,k=1}^m u_{jk} = l-1, \, \sum_{j=1}^m u_{jj} = 0, \, \sum_{j=1}^m u_{ji} \leq 1 \, &&(\forall i \in \{2,\cdots,m\}) \\
	& v_j - v_k + 1 \le (m-1) (1-u_{jk}) \, &&(\forall j,k \in \{2,\cdots,m\}).
	\end{align*}
\end{minipage}
}

\vspace{0.5\baselineskip}

Here, we index POIs such that $s = p_1$ for brevity.
The binary $u_{jk}$ are true iff
we visit $p_k$ immediately after visiting $p_j$;
the integer $v_j$ track the rank of $p_j$;
and $z(\bu)_i \defEq \sum_{j=1}^m u_{ji} - \sum_{k=1}^m u_{ik}$ indicates whether we end up at $p_i$.
The constraints ensure we output a path of exactly $l$ POIs;
the last constraint in particular ensures we do not have (disjoint) cycles, as per \citet{Miller:1960}.
By reading off the values of $u_{jk}$, we can determine the predicted sequence.

An off-the-shelf solver (\eg {\tt Gurobi}) may be used on the ILP.
While ILPs have worst case exponential complexity,
such solvers are highly optimised and thus make many problem instances tractable.

The above ILP implicitly solves both the problem of selecting the set of $l$ POIs to recommend,
and the problem of ordering them.
We could fix the set of POIs to recommend, \eg by using the POIs in the Viterbi solution, and then find the optimal ordering of these nodes.
However, we have found this to have only modest improvement over the loop elimination heuristic of the previous section.

\section{Top-k Sequences using List Viterbi}
\label{sec:viterbi}


The Viterbi algorithm 
finds the best scoring sequence in Equation \ref{eqn:argmax}, which may have loops.
To find the best sequence without loops in Equation \ref{eqn:argmax-path}, one can apply a \emph{list Viterbi algorithm}
to find not just the single best sequence,
but rather the top $K$ best sequences.
By definition, the first such sequence that is loop free must be the highest scoring sequence that does not have loops.
We detail existing approaches to solve the list Viterbi problem.

\subsection{Parallel and serial list Viterbi algorithms}

At a high level, there are two approaches to extend the Viterbi algorithm to the top-$K$ setting.
The first approach is to keep track, at each state, of the top $K$ sub-sequences that end at this state; these are known as \emph{parallel list Viterbi} algorithms.
Such algorithms date back to at least \citet{Forney:1973},
but have the disadvantage of imposing a non-trivial memory burden.
Further, they are not applicable as-is in our setting:
we do \emph{not} know in advance what value of $K$ is suitable,
since we do not know the position of the best loop-free sequence.

The second approach is to more fundamentally modify how one selects paths; these are known as \emph{serial list Viterbi} algorithms.
There are at least two such well-known proposals in the context of hidden Markov models (HMMs).
In the signal processing community, \citet{seshadri1994list} proposed an algorithm that keeps track of the ``next-best'' sequence terminating at each state in the current list of best sequences.
In the AI community, \citet{nilsson2001sequentially}
proposed an algorithm that cleverly partitions the search space into subsets of sequences that share a prefix with the current list of best sequences.
While derived in different communities, these two algorithms are in fact only superficially different, as we now see.

\subsection{Relating serial list Viterbi algorithms}

The connection between the two list Viterbi algorithms is easiest to see when finding the second-best sequence for an HMM.
Suppose we have an HMM with states $\SSf_{t}$, observations $\OSf_{t}$, transitions $a(i, j) = \Pr( \SSf_{t+1} = j \mid \SSf_t = i )$, and emissions $b(i, k) = \Pr( \OSf_{t} = k \mid \SSf_t = i )$.
Suppose $s^*_{1:T}$ is the most likely length $T$ sequence given observations $\OSf_{1:T}$, and
$\delta(j, t)$ is the value of the best sequence up till position $t$ ending at state $j$ as computed by the Viterbi algorithm.

Our interest is in finding the second-best sequence 
with value
$ M \defEq \max_{\SSf_{1:T} \neq s^*_{1 : T}}{\Pr( \SSf_{1:T}, \OSf_{1:T} )}. $
\citet{seshadri1994list} 
observed that $M = \bar{\delta}_{T+1}$, where $\bar{\delta}_t$ has a Viterbi-like recurrence
\begin{equation}
    \label{eqn:att-recurrence}
    \resizebox{0.9\linewidth}{!}{$
    \begin{aligned}
        \bar{\delta}_t &\defEq 
        \indicator{t > 0} \cdot
        \max
        \begin{cases}
        \max_{i \neq s^*_{t-1}} \delta(i, t-1) \cdot a( i, s^*_{t} ) \cdot b( s^*_{t}, \OSf_{t} ) \\
        \bar{\delta}_{t - 1} \cdot a( s^*_{t - 1}, s^*_{t} ) \cdot b( s^*_{t}, \OSf_{t} ).
        \end{cases}
    \end{aligned}
    $}
\end{equation}
Intuitively, $\bar{\delta}_t$ finds the value of the second-best sequence that merges with the best sequence by at least time $t$.

\citet{nilsson2001sequentially} observed that $M = \max_t \widehat{\rho}_t$, where
\begin{align*}
	\widehat{\rho}_{t} &\defEq \max_{i \neq s^*_{t}} \max_{S_{t+1:T}} {\Pr( \SSf_{1:t-1} = s^*_{1:t-1}, \SSf_t = i, \SSf_{t+1:T}, \OSf_{1:T} )}.
\end{align*}
Intuitively, $\widehat{\rho}_t$ finds the value of the second-best sequence that first deviates from the best sequence exactly at time $t$.
One can compute $\widehat{\rho}_{t}$ using $\eta_{i, j, t} \defEq \max_{\SSf \colon S_{t - 1} = i, \SSf_{t} = j} \Pr( \SSf_{1:T}, \OSf_{1:T} )$ \citep{nilsson2001sequentially},
which in turn can be computed from the ``backward'' analogue of $\delta$.

To connect the two approaches, by unrolling the recurrence in Equation \ref{eqn:att-recurrence}, and by definition of $\delta$, we have
$M = \max_t \widehat{\mu}_t$ where
\resizebox{\linewidth}{!}{
    \begin{minipage}{\linewidth}
        \begin{align*}
        	\widehat{\mu}_{t} &\defEq \left[ \prod_{k = t+2}^T a( s^*_{k-1}, s^*_{k} ) \cdot b( s^*_{k}, \OSf_{k} ) \right] \cdot \max_{i \neq s^*_{t}} \delta(i, t) \cdot a( i, s^*_{t+1} ) \cdot b( s^*_{t+1}, \OSf_{t+1} ) \\
        	&= \max_{i \neq s^*_{t}} \max_{S_{1:t-1}} {\Pr( \SSf_{1:t-1}, \SSf_t = i, \SSf_{t+1:T} = s^*_{t+1:T}, \OSf_{1:T} )};
        \end{align*}
    \end{minipage}
}%

\vspace{0.5\baselineskip}
\noindent i.e., the same quantities are computed, except that the former fixes the suffix of the candidate sequence, while the latter fixes the prefix.
A similar analysis holds in the case of finding the $K$th best sequence.

The complexity of either of the above algorithms is $\mathscr{O}( l \cdot |\PCal|^2 + l \cdot |\PCal| \cdot K + l \cdot K \cdot \log( l \cdot K ) )$ \citep{nilsson2001sequentially}.
This is tractable, but we emphasise that the smallest $K$ guaranteeing we obtain a path is unknown \emph{a-priori}.

\section{Empirical comparison}
\label{sec:experiments}

We now empirically compare the methods discussed above,
to get a firmer sense of their tradeoffs.
We focus on path recommendation tasks where an SSVM is learned to score sequences.
(We refer the reader to \citet{Chen:2017} for a detailed comparison of SSVM to other baselines, e.g. RankSVM.)
We then apply 
each of the above methods to (approximately) solve the inference problem of Equation \ref{eqn:argmax-path}.

%
\subsection{Experimental setup}

Following \citet{cikm16paper,Chen:2017},
we work with
data
extracted from Flickr photos for the cities of {\tt Glasgow}, {\tt Osaka} and
{\tt Toronto}~\cite{ijcai15,cikm16paper}.
Each dataset comprises of a
list of trajectories as visited by various Flickr users. 
Table~\ref{tab:data} summarises the statistics of each dataset.

\begin{table}[t]
		\setlength{\tabcolsep}{4pt} 
		\small
		\begin{tabular}{llll} \hline 
		\textbf{Dataset} & \textbf{\# Traj} & \textbf{\# POIs}  & \textbf{\# Queries} \\ \hline
		{\tt Glasgow}          & 351              & 25              & 64 \\
		{\tt Osaka}            & 186              & 26              & 47 \\
        {\tt Toronto}          & 977              & 27              & 99 \\
		\hline
		\end{tabular}%
		\captionof{table}{Statistics of trajectory datasets: the number of trajectories (\# Traj), POIs (\# POIs), queries (\# Queries). Note that a distinct query may be associated with multiple trajectories.}
		\label{tab:data}
	\vspace{-2\baselineskip}
\end{table}

%

To produce a recommendation,
we use 
the aforementioned inference methods (named {\sc LoopElim(++)}, {\sc Greedy}, {\sc ILP}, {\sc ListViterbi})
as well as
standard inference ({\sc Viterbi}).
{\sc LoopElim} processes the Viterbi solution for the query length $l$,
while {\sc LoopElim++} processes the solutions for all longer lengths $l'$, as described in \S\ref{sec:loop-elim}.

We evaluate each algorithm using leave-one-query-out cross validation,
\ie in each round, we hold out all trajectories for a distinct query $\x$ in the dataset.
To measure performance,
we use 
the {\bf F$_1$ score on points}~\cite{ijcai15}, which computes F$_1$-score on the predicted versus seen points
without considering their relative order,
and the {\bf F$_1$ score on pairs}~\cite{cikm16paper}, which computes the F$_1$-score on all ordered pairs in the predicted versus ground truth sequence. 


\begin{table*}[t]
	\caption{Point F$_1$ cross-validation scores.}
	\label{tab:f1-master}
	\centering
	\vspace{-\baselineskip}
	\resizebox{0.575\textwidth}{!}{
	\subfloat[Raw scores.]{		
	\begin{tabular}{l|ccccc|c}
		\hline
         & \textsc{LoopElim} & \textsc{LoopElim++} & \textsc{Greedy} & \textsc{ILP} & \textsc{ListViterbi} & \textsc{Viterbi} \\ \hline
        Osaka & $0.635\pm0.039$ & $\cellcolor{gray!25}{0.639\pm0.038}$ & $0.626\pm0.040$ & $0.638\pm0.039$ & $0.638\pm0.039$ & $0.622\pm0.040$ \\
        Glasgow & $0.718\pm0.030$ & $0.721\pm0.030$ & $\cellcolor{gray!25}{0.751\pm0.028}$ & $0.741\pm0.028$ & $0.741\pm0.028$ & $0.689\pm0.032$ \\
        Toronto & $0.699\pm0.027$ & $0.696\pm0.027$ & $\cellcolor{gray!25}{0.756\pm0.024}$ & $0.754\pm0.023$ & $0.754\pm0.023$ & $0.651\pm0.030$ \\
		\hline
	\end{tabular}
	}
	\label{tab:f1}%
	}%
	\quad	
	\resizebox{0.4\textwidth}{!}{
	\subfloat[Improvement over \textsc{Viterbi}.]{
	\begin{tabular}{l|ccccc}
		\hline
         & \textsc{LoopElim} & \textsc{LoopElim++} & \textsc{Greedy} & \textsc{ILP} & \textsc{ListViterbi} \\ \hline
        Osaka & $2.0\%$ & $\cellcolor{gray!25}{2.7\%}$ & $0.5\%$ & $2.6\%$ & $2.6\%$ \\
        Glasgow & $4.2\%$ & $4.6\%$ & $\cellcolor{gray!25}{9.0\%}$ & $7.6\%$ & $7.6\%$ \\
        Toronto & $7.3\%$ & $6.8\%$ & $\cellcolor{gray!25}{16.0\%}$ & $15.7\%$ & $15.7\%$ \\
		\hline
	\end{tabular}
	\label{tab:f1-up}%
	}
	}
\end{table*}

\begin{table*}[t]
	\caption{Pair F$_1$ cross-validation scores.}
	\label{tab:pf1-master}	
	\centering
	\vspace{-\baselineskip}
	\resizebox{0.575\textwidth}{!}{
	\subfloat[Raw scores.]{
	\label{tab:pf1}
	\begin{tabular}{l|ccccc|c}
		\hline
         & \textsc{LoopElim} & \textsc{LoopElim++} & \textsc{Greedy} & \textsc{ILP} & \textsc{ListViterbi} & \textsc{Viterbi} \\ \hline
        Osaka & $0.369\pm0.059$ & $0.373\pm0.059$ & $0.369\pm0.059$ & $\cellcolor{gray!25}{0.375\pm0.059}$ & $\cellcolor{gray!25}{0.375\pm0.059}$ & $0.364\pm0.060$ \\
        Glasgow & $0.480\pm0.049$ & $0.485\pm0.049$ & $\cellcolor{gray!25}{0.522\pm0.048}$ & $0.506\pm0.048$ & $0.508\pm0.048$ & $0.461\pm0.050$ \\
        Toronto & $0.490\pm0.041$ & $0.489\pm0.041$ & $\cellcolor{gray!25}{0.543\pm0.038}$ & $0.530\pm0.037$ & $0.529\pm0.037$ & $0.463\pm0.041$ \\
		\hline
	\end{tabular}
	}
	}%
	\quad
	\resizebox{0.4\textwidth}{!}{
	\subfloat[Improvement over \textsc{Viterbi}.]{
		\label{tab:pf1-up}
		\begin{tabular}{l|ccccc}
			\hline
             & \textsc{LoopElim} & \textsc{LoopElim++} & \textsc{Greedy} & \textsc{ILP} & \textsc{ListViterbi} \\ \hline
            Osaka & $1.4\%$ & $2.4\%$ & $1.4\%$ & $\cellcolor{gray!25}{3.0\%}$ & $\cellcolor{gray!25}{3.0\%}$ \\
            Glasgow & $4.1\%$ & $5.1\%$ & $\cellcolor{gray!25}{13.1\%}$ & $9.7\%$ & $10.1\%$ \\
            Toronto & $5.8\%$ & $5.7\%$ & $\cellcolor{gray!25}{17.2\%}$ & $14.5\%$ & $14.4\%$ \\
			\hline
		\end{tabular}
	}
	}
\end{table*}

\begin{table*}[t]
    \caption{Point F$_1$ cross-validation scores of queries where {\sc Viterbi} recommendation has a loop.}
    \label{tab:f1-diff}
	\centering	
	\vspace{-\baselineskip}
	\resizebox{0.575\textwidth}{!}{
    \subfloat[Raw scores.]{
	\begin{tabular}{l|ccccc|c}
		\hline
         & \textsc{LoopElim} & \textsc{LoopElim++} & \textsc{Greedy} & \textsc{ILP} & \textsc{ListViterbi} & \textsc{Viterbi} \\ \hline
        Osaka & $0.389\pm0.038$ & $\cellcolor{gray!25}{0.408\pm0.038}$ & $0.374\pm0.046$ & $0.405\pm0.042$ & $0.405\pm0.042$ & $0.338\pm0.034$ \\
        Glasgow & $0.569\pm0.045$ & $0.577\pm0.045$ & $\cellcolor{gray!25}{0.658\pm0.044}$ & $0.644\pm0.040$ & $0.644\pm0.040$ & $0.476\pm0.040$ \\
        Toronto & $0.540\pm0.030$ & $0.534\pm0.030$ & $\cellcolor{gray!25}{0.657\pm0.027}$ & $0.654\pm0.025$ & $0.654\pm0.025$ & $0.443\pm0.028$ \\
		\hline
	\end{tabular}
	}
    \label{tab:f1-diff}
	}%
	\quad
	\resizebox{0.4\textwidth}{!}{
	\subfloat[Improvement over \textsc{Viterbi}.]{
	\begin{tabular}{l|ccccc}
		\hline
         & \textsc{LoopElim} & \textsc{LoopElim++} & \textsc{Greedy} & \textsc{ILP} & \textsc{ListViterbi} \\ \hline
        Osaka & $15.1\%$ & $\cellcolor{gray!25}{20.9\%}$ & $10.8\%$ & $19.7\%$ & $19.7\%$ \\
        Glasgow & $19.4\%$ & $21.3\%$ & $\cellcolor{gray!25}{38.2\%}$ & $35.2\%$ & $35.2\%$ \\
        Toronto & $22.1\%$ & $20.7\%$ & $\cellcolor{gray!25}{48.5\%}$ & $47.8\%$ & $47.7\%$ \\
		\hline
	\end{tabular}
	\label{tab:f1-diff-up}%
	}
	}
\end{table*}

\begin{table*}[t]
	\caption{Pair F$_1$ cross-validation scores of queries where {\sc Viterbi} recommendation has a loop.}
	\label{tab:pf1-diff}
	\centering
	\vspace{-\baselineskip}
	\resizebox{0.575\textwidth}{!}{
	\subfloat[Raw scores.]{
	\label{tab:pf1-diff}
	\begin{tabular}{l|ccccc|c}
		\hline
         & \textsc{LoopElim} & \textsc{LoopElim++} & \textsc{Greedy} & \textsc{ILP} & \textsc{ListViterbi} & \textsc{Viterbi} \\ \hline
        Osaka & $0.088\pm0.026$ & $0.104\pm0.026$ & $0.103\pm0.034$ & $\cellcolor{gray!25}{0.112\pm0.030}$ & $\cellcolor{gray!25}{0.112\pm0.030}$ & $0.067\pm0.020$ \\
        Glasgow & $0.261\pm0.047$ & $0.277\pm0.052$ & $\cellcolor{gray!25}{0.378\pm0.066}$ & $0.345\pm0.055$ & $0.350\pm0.054$ & $0.201\pm0.039$ \\
        Toronto & $0.236\pm0.031$ & $0.235\pm0.030$ & $\cellcolor{gray!25}{0.352\pm0.032}$ & $0.318\pm0.024$ & $0.317\pm0.023$ & $0.180\pm0.025$ \\
		\hline
	\end{tabular}
	}
	}%
	\quad
	\resizebox{0.4\textwidth}{!}{
	\subfloat[Improvement over \textsc{Viterbi}.]{
		\label{tab:pf1-diff-up}
		\begin{tabular}{l|ccccc}
			\hline
             & \textsc{LoopElim} & \textsc{LoopElim++} & \textsc{Greedy} & \textsc{ILP} & \textsc{ListViterbi} \\ \hline
            Osaka & $32.7\%$ & $55.4\%$ & $54.5\%$ & $\cellcolor{gray!25}{68.2\%}$ & $\cellcolor{gray!25}{68.2\%}$ \\
            Glasgow & $29.9\%$ & $37.6\%$ & $\cellcolor{gray!25}{88.1\%}$ & $71.4\%$ & $73.9\%$ \\
            Toronto & $30.9\%$ & $30.1\%$ & $\cellcolor{gray!25}{95.0\%}$ & $76.5\%$ & $76.0\%$ \\
			\hline
		\end{tabular}
	}
	}
\end{table*}

\subsection{Results and discussion}

We now address several key questions relating to the tradeoffs of the various methods considered.

\textbf{How often does the top-scoring sequence have loops?}
It is first of interest to confirm that
even for the powerful SSVM model,
the top-scoring sequence as found by the Viterbi algorithm often contain loops.
Indeed, we find that on the ({\tt Osaka}, {\tt Glasgow}, {\tt Toronto}) datasets, the top-scoring sequence for (23.9\%, 31.2\%, 48.5\%) respectively of all queries have loops.

\textbf{How important is it to remove loops?}
Having confirmed that loops in the top-scoring sequence are an issue,
it is now of interest to establish that removing such loops during prediction is in fact important.
This is confirmed in Tables \ref{tab:f1-master} -- \ref{tab:pf1-master},
where we see that there can be as much as a \textbf{17\%} improvement in performance over the {\sc Viterbi} baseline.
These improvements are over all queries, including those where the {\sc Viterbi} algorithm does not have loops.
Restricting to those queries where there are loops,
Tables \ref{tab:f1-diff-up} -- \ref{tab:pf1-diff-up}
show that the improvements are dramatic, being as high as \textbf{50\%}.

\begin{figure*}[!t]
		\quad
		\centering
		\includegraphics[width=0.95\textwidth]{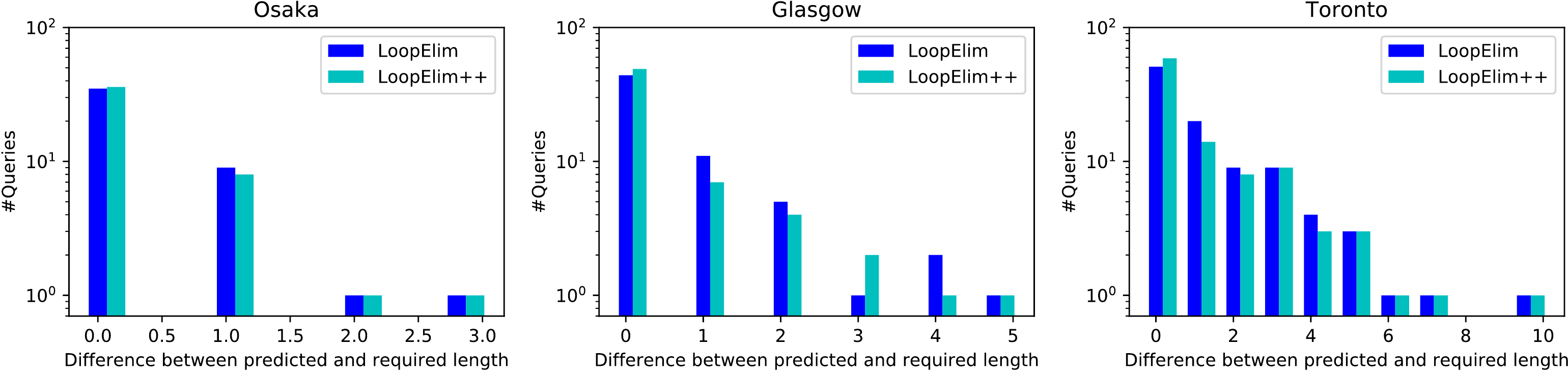}
	    \captionof{figure}{Absolute difference between recommended and required sequence length for {\sc LoopElim(++)}.}
	    \label{fig:length-christo}
\end{figure*}

\textbf{How reliably can {\sc LoopElim(++)} get the desired length?}
Recall that
{\sc LoopElim} and its variant {\sc LoopElim++}
may result in a path of the wrong length.
Figure \ref{fig:length-christo} shows that for a significant fraction of queries,
these algorithms
will output a different length path to that specified in the query,
and are thus not suitable if we strictly enforce a length constraint.
Of the two, {\sc LoopElim++} outputs more trajectories of the correct length, as per design.

\textbf{How reliably can {\sc LoopElim(++)} predict a good path?}
Assuming one can overlook {\sc LoopElim(++)} producing a path of possibly incorrect length,
it is of interest as to how well
they perform.
Tables \ref{tab:f1-master} -- \ref{tab:pf1-master} show that the heuristic often
grossly underperforms compared to the exact {\sc ILP} and {\sc ListViterbi} approaches.
(Being exact, the latter methods have nearly identical accuracy, with occasional differences owing to ties.)
Curiously, {\sc LoopElim++}, while producing paths of length closer to the original, actually performs slightly \emph{worse} than the na\"{i}ve {\sc LoopElim} on the larger {\tt Toronto} dataset.


\textbf{How reliably can {\sc Greedy} predict a good path?}
Unlike {\sc LoopElim(++)}, the {\sc Greedy} method is guaranteed to produce a path of the correct length.
Surprisingly, it also performs very well compared to the {\sc ListViterbi} and {\sc ILP} methods, 
offering significant improvements over these methods on the larger {\tt Glasgow} and {\tt Toronto} datasets,
while being competitive on the smaller {\tt Osaka} dataset.

\textbf{How does trajectory length influence accuracy?}
The above analyses the accuracy over all queries, and over queries where {\sc Viterbi} outputs a loop.
It is of interest to partition the set of queries based on the length of the requested trajectory.
Intuitively, we expect that the longer the requested trajectory, the less accurate all methods will fare; this is because longer trajectories imply an 
exponentially large search space.
(Indeed, on {\tt Toronto}, for a query with length 13, the first \textbf{5 million} sequences have loops!)

\begin{figure*}[!t]
		\quad
		\centering
		\includegraphics[width=0.95\textwidth]{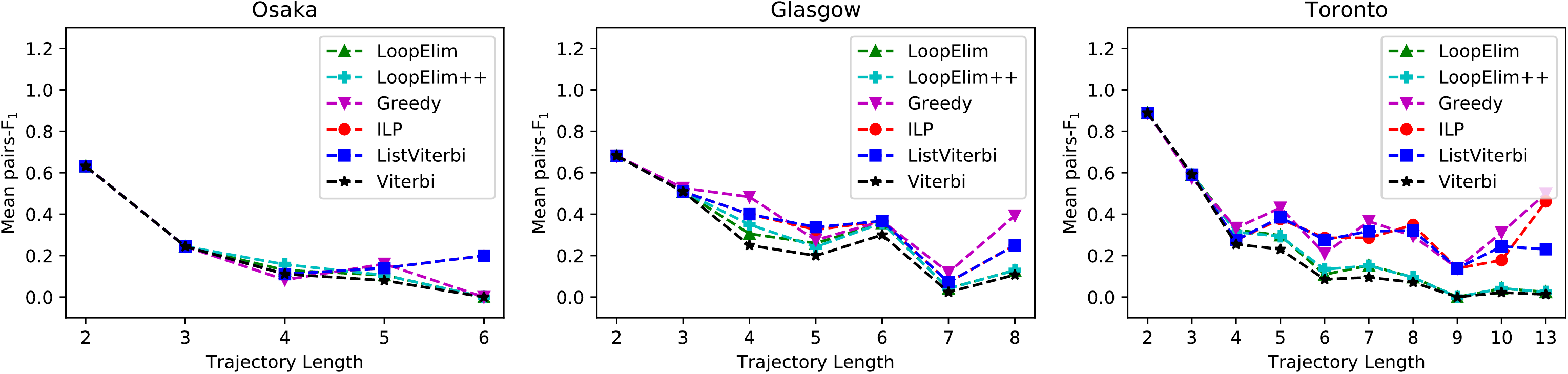}
	    \captionof{figure}{Accuracy versus trajectory length for all inference algorithms.}
	    \label{fig:acc-vs-length}
\end{figure*}

Figure \ref{fig:acc-vs-length} confirms this intuition:
on all datasets, and for all methods,
a longer trajectory length implies significantly worse performance in absolute terms.
Interestingly, the relative improvements of all methods over {\sc Viterbi} are either consistent or actually \emph{increase} with longer trajectories;
this is reassuring, and justifies the effort spent in removing loops.
Of further interest is that the {\sc Greedy} heuristic remains dominant for longer trajectories on the {\tt Glasgow} and {\tt Toronto} datasets.

\begin{figure*}[!t]
		\centering
		\includegraphics[width=0.95\textwidth]{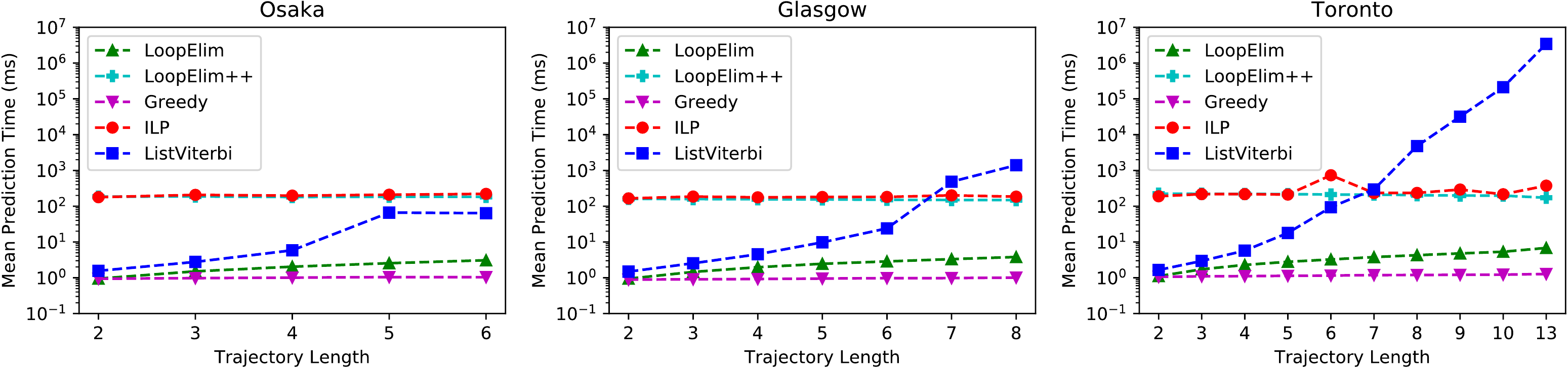}
	    \captionof{figure}{Prediction time versus trajectory length for all inference algorithms.}
	    \label{fig:inftime}
\end{figure*}

\textbf{How fast are the various methods?}
As expected, the heuristic {\sc LoopElim} and {\sc Greedy} algorithms have the fastest runtime, being on the order of milliseconds per query even for long trajectories (Figure \ref{fig:inftime}).
The {\sc Greedy} algorithm is the faster of the two, as it does not even require running the standard Viterbi algorithm.
The {\sc LoopElim++} variant is much slower than {\sc LoopElim}, as it needs to perform the Viterbi calculation multiple times.

The exact methods are by comparison slower, especially for medium length trajectories.
Amongst these methods, for shorter trajectories, the {\sc ListViterbi} approach is to be preferred;
however, for longer trajectories, the {\sc ILP} approach is faster.
The reason for the {\sc ListViterbi} to suffer at longer trajectories is simply because this creates an exponential increase in the number of available choices, which must be searched through serially.
Of interest is that {\sc ILP} approach has runtime largely independent of the trajectory length.
This indicates the branch-and-bound as well as cutting plane underpinnings of these solvers are highly scalable.

Overall, the {\sc Greedy} algorithm is at least competitive, and often more accurate than exact methods;
it is also significantly faster.
Thus, for recommending paths, we recommend this algorithm.




\section{Discussion and Conclusion}
\label{sec:discussion}

We formalised the problem of eliminating loops when recommending trajectories to visitors in a city,
and surveyed three distinct approaches to the problem --
graph-based heuristics,
list extensions of the Viterbi algorithm,
and integer linear programming.
We explicated how two ostensibly different approaches to the list Viterbi algorithm \citep{seshadri1994list,nilsson2001sequentially} are in fact fundamentally identical.

In experiments on real-world datasets,
a greedy graph-based heuristic offered excellent performance and runtime.
We thus recommend its use
for removing loops at prediction time
over the more involved integer programming and list Viterbi algorithms.

As a caveat on the applicability of the greedy algorithm,
we note that the problem of removing loops also arises during training SSVMs in the loss-augmented inference step \citep{Chen:2017}.
The list Viterbi algorithm has been demonstrated useful in this context;
it is unclear whether the same will be true of the approximate greedy algorithm, as it will necessarily lead to sub-optimal solutions.

As future work, it is of interest to extend the greedy algorithm to the top-$K$ evaluation setting of \citet{Chen:2017}, wherein the recommender produces a \emph{list} of paths to be considered.
A natural strategy would be to augment the algorithm with a beam search.

Further, the idea of modifying the standard Viterbi inference problem (Equation \ref{eqn:argmax}) has other applications, such
as ensuring diversity in the predicted ranking.
Such problems have been studied in contexts such as information retrieval \citep{Carbonell:1998} and computer vision \citep{Park:2011},
and their study would be interesting in trajectory recommendation.
More broadly, investigation of efficient means of ensuring global cohesion -- \eg preventing homogeneous results -- 
is an important direction for the advancement of citizen-centric recommendation.



\end{document}